\def\year{2020}\relax
\documentclass[letterpaper]{article} 
\usepackage{aaai20}  
\usepackage{times}  
\usepackage{helvet} 
\usepackage{courier}  
\usepackage[hyphens]{url}  
\usepackage{graphicx} 
\usepackage{graphicx}  
\usepackage{latexsym}
\usepackage{multirow}
\usepackage{colortbl}
\usepackage{booktabs}
\usepackage{threeparttable}

\usepackage{amsmath}
\usepackage{amssymb}
\usepackage{amsfonts,amsthm}

\usepackage{setspace}

\newtheoremstyle{mytask}%
{}{}{\normalfont}{}{\bf}{ }{ }{}
\theoremstyle{mytask}
\newtheorem{task}{Task}

\newcommand{\citet}[1]{\citeauthor{#1} \shortcite{#1}} \newcommand{\citep}{\cite} 

\nocopyright
\title{Length-controllable Abstractive Summarization \\ by Guiding with Summary Prototype}
\author{}
\author{Itsumi Saito\textsuperscript{\rm 1,2} , Kyosuke Nishida\textsuperscript{\rm 1}, Kosuke Nishida\textsuperscript{\rm 1}, Atsushi Otsuka\textsuperscript{\rm 1}, Hisako Asano\textsuperscript{\rm 1}, \\ \Large \bf Junji Tomita\textsuperscript{\rm 1}, \bf Hiroyuki Shindo\textsuperscript{\rm 2}, \bf Yuji Matsumoto\textsuperscript{\rm 2,3}\\ 
NTT Media Intelligence Laboratories, NTT Corporation\textsuperscript{\rm 1}\\ 
Nara Institute of Science and Technology\textsuperscript{\rm 2}\\
RIKEN Center for Advanced Intelligence Project\textsuperscript{\rm 3} \\
itsumi.saito.df@hco.ntt.co.jp 
}

\begin{document}

\maketitle

\begin{abstract}
We propose a new length-controllable abstractive summarization model. 
Recent state-of-the-art abstractive summarization models based on 
encoder-decoder models generate only one summary per source text. However, controllable summarization, especially of the length, is an important aspect for practical applications. 
Previous studies on length-controllable abstractive summarization incorporate length embeddings in the decoder module for controlling the summary length. Although the length embeddings can control 
where to stop decoding, 
\textcolor{black}{they do not decide which information should be included in the summary within the length constraint.}
Unlike the previous models, \textcolor{black}{our length-controllable abstractive summarization model incorporates} a word-level extractive module 
in the encoder-decoder model instead of length embeddings. 
Our model generates a summary in two steps. First, our word-level extractor extracts 
a sequence of important words (we call it the ``prototype text") from the source text according to the word-level importance scores and the length constraint. Second, the prototype text is used as additional input to the encoder-decoder model, which generates a summary 
\textcolor{black}{by jointly encoding and copying words from} 
both the prototype text and source text. 
Since the prototype text is a guide to both 
\textcolor{black}{the content and length of the summary}, our model can generate an 
informative and length-controlled summary. Experiments with the CNN/Daily Mail dataset and the NEWSROOM dataset show that our model outperformed previous models in length-controlled settings.
\end{abstract}

\section{Introduction}
\begin{figure}[t!]
\begin{center}
\begingroup
\begin{spacing}{0.7}
{
\tabcolsep=5pt
\begin{tabular}{|p{23em}|}
\hline
\\[-.5em]
{\bf {Source Text}}  \\
\\[-.5em]
{\small 
various types of renewable energy such as solar and wind are often touted as being the solution to the world 's growing energy crisis . but one researcher has come up with a novel idea that could trump them all - a biological solar panel that works around the clock . by harnessing the electrons generated by plants such as moss , he said he can create useful energy that could be used at home or elsewhere . \textcolor{black}{a \textcolor{blue}{university of cambridge scientist has revealed his} green source of energy . by using just moss he is able to generate enough power to run a} clock -lrb- shown -rrb- . \textcolor{black}{\textcolor{blue}{\textcolor{red}{\bf he said panels of plant material could power} appliances \textcolor{red}{in our} homes}} . \textcolor{black}{\textcolor{blue}{and the technology could help farmers grow crops where} electricity is scarce} . (...)
} 
\\
\\[-.3em]
{\bf {Reference Summary}} \\
\\[-.5em]
{\small university of cambridge scientist has revealed his green source of energy . by using just moss he is able to generate enough power to run a clock . he said panels of plant material could power appliances in our homes . and the tech could help farmers grow crops where electricity is scarce . }\\
\\[-.3em]
{\bf Outputs (K=10)} \\
\\[-.3em]
{\small [{\bf Extracted prototype}] he said panels of plant material could power in our } \\
{\small [{\bf Abstractive summary}]  panels of plant material could power appliances . } \\
\\[-.3em]
{\bf Outputs (K=30)} \\
\\[-.3em]
{\small [{\bf Extracted Prototype}] university of cambridge scientist has revealed his he said panels of plant material could power appliances in our homes and the technology could help farmers grow crops where is scarce } \\
{\small [{\bf Abstractive summary}] university of cambridge scientist has revealed his green source of energy . he said panels of plant material could power appliances in our homes .} \\[-.5em]
\\
\hline
\end{tabular}
}
\end{spacing}
\endgroup
\end{center}
\caption{\label{fig:example_outputs} Output examples of our model. Our model extracts the 
top-$K$ important words, which are colored red ($K=10$) and blue ($K=30$),
as a prototype 
from the source text.
It generates an abstractive summary based on the prototype and source texts. The length of the generated summary is controlled in accordance with the length of the prototype text.}
\end{figure}

Neural summarization has made great progress in recent years. It has two main approaches: extractive and abstractive. Extractive methods generate summaries by selecting important sentences~\cite{latent_select,score_and_select}. They produce grammatically correct summaries; however, they do not give much flexibility to the summarization because they only extract sentences from the source text. 
By contrast, abstractive summarization enables more flexible summarization, and it is expected to generate more fluent and readable summaries than extractive models. The most commonly used abstractive summarization model is the pointer-generator~\cite{see17}, which generates a summary word-by-word while copying words from the source text and generating words from a pre-defined vocabulary set. This model can generate an accurate summary by combining word-level extraction and generation.

Although the idea of controlling the length of the summary was mostly neglected in the past, it was recently pointed out that it is actually an important aspect of abstractive summarization~\cite{CNNlengthcontrol,controllable_summarization}.  In practical applications, the summary length 
should be controllable \textcolor{black}{in order for it} to fit the device that displays it.
However, there have only been a few studies on controlling the summary length. \citet{kikuchi16} 
proposed a length-controllable model that uses length embeddings.
In the length embedding approach, the summary length is 
\textcolor{black}{encoded either as an embedding that represents the remaining length at each decoding step
or as an initial embedding to the decoder that represents the 
\textcolor{black}{desired length}.
} 
\citet{CNNlengthcontrol} proposed a 
model 
\textcolor{black}{that uses the desired length as an input to the initial state of the decoder.} 
These previous models control the length in the decoding module 
\textcolor{black}{by} 
using length embeddings. However, length embeddings only add length information on the decoder side.  Consequently, they may miss important information because it is difficult to take into account which \textcolor{black}{content} should be included in the summary for certain length constraints.

\begin{figure}[t]
\centering
 \includegraphics[width=8.3cm]{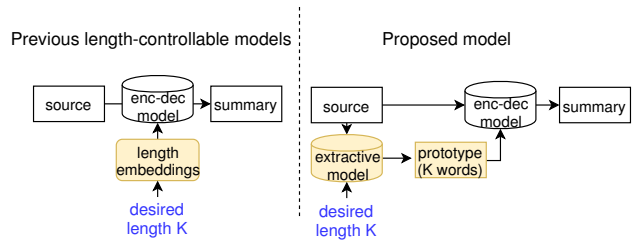}
  \caption{Comparison of previous length-controllable models and proposed model. Our model 
  controls the summary length in accordance with the length of the prototype text.
  \label{fig:compare_lc}}
\end{figure}

We propose a new length-controllable abstractive summarization \textcolor{black}{that is guided by} the prototype text. 
Our idea is to 
\textcolor{black}{use} 
a word-level extractive module instead of length embeddings to control the summary length. \textcolor{black}{Figure~\ref{fig:compare_lc} compares the previous length-controllable models and the proposed 
\textcolor{black}{one}. 
The yellow blocks are the modules responsible for length control.}
Since the word-level extractor controls which \textcolor{black}{contents} are to be included in the summary when a length constraint is given, it is possible to generate a summary including the important contents. Our model consists of two steps. First, the word-level extractor predicts the word-level importance of the source text and extracts important words according to the 
importance scores and the desired length. The extracted word sequence is used as a ``prototype" of the summary; we call it the “prototype” text. Second, we use the prototype text as an additional input of the encoder-decoder model. 
\textcolor{black}{The length of the summary is kept close to that of the prototype text because the summary is generated by referring to the prototype text.}
\textcolor{black}{Figure~\ref{fig:example_outputs} shows examples of output generated by our model. Our abstractive summaries are similar to the extracted prototypes. The extractive module produces a rough overview of the summary, and the encoder-decoder module produces a fluent summary based on the extracted prototype.}

\textcolor{black}{Our idea is inspired by extractive-and-abstractive summarization. Extractive-and-abstractive summarization incorporates an extractive model in an abstractive \textcolor{black}{encoder-decoder} model. 
\textcolor{black}{While} in the simple encoder-decoder model, 
\textcolor{black}{one model identifies the }
important contents and generates fluent summaries, 
the extractive-and-abstractive model
\textcolor{black}{has an }
encoder-decoder 
\textcolor{black}{part that }
generates fluent summaries 
and a 
\textcolor{black}{separate part that extracts }
important contents. 
Several studies have shown that separating the problem of finding the important content and 
the problem of generating fluent summaries improves the accuracy of the summary~\cite{bottomup,sentence_rewriting}. 
Our model can be regarded as an extension of 
\textcolor{black}{models that work in this way: }
However, this is the first to 
\textcolor{black}{extend the extractive module such that it can control the summary length. }
}

\textcolor{black}{Ours}
is the first method that controls the summary length using an extractive module and that achieves both high accuracy and length controllability in abstractive summarization. 
Our contributions are summarized as follows:
\begin{itemize}
\item We propose a new length-controllable prototype-guided abstractive summarization model, called LPAS \textcolor{black}{(Length-controllable Prototype-guided Abstractive Summarization). }
Our model 
effectively guides the abstractive summarization 
\textcolor{black}{using a summary prototype}. 
Our model controls the summary length 
by controlling the number of words in the prototype text.
\item \textcolor{black}{Our model achieved state-of-the-art ROUGE scores in length-controlled abstractive summarization settings on the CNN/DM and NEWSROOM datasets.}
\end{itemize}

\section{Task Definition}
Our study 
\textcolor{black}{defines length-controllable abstractive summarization as two pipelined tasks: 
prototype extraction and 
prototype-guided abstractive summarization}.
The problem formulations of each task are described below.

\begin{task}
[Prototype Extraction] Given a source text $X^C$ with $L$ words $X^C$= $(x^C_1,\dots,x^C_L)$ \textcolor{black}{and a  desired summary length $K$}, \textcolor{black}{the model} 
\textcolor{black}{estimates importance scores $P^{\rm ext}$= $(p^{\rm ext}_1 \ldots p^{\rm ext}_L)$ and} 
extracts \textcolor{black}{the top-$K$ \textcolor{black}{important} words $X^P$= $(x^P_1,\ldots,x^P_K)$ as} a prototype text on the basis of $P^{\rm ext}$. 
The desired summary length $K$ can be set to an arbitrary value. 
\textcolor{black}{Note that the original word order is preserved in $X^{P}$ ($X^{P}$ is not bag-of-words).}
\end{task}
\begin{task}
[\textcolor{black}{Prototype-guided} Abstractive Summarization] Given \textcolor{black}{the} source text 
and \textcolor{black}{the extracted prototype text $X^{P}$}, \textcolor{black}{the model} generates a length-controlled abstractive summary $Y$ = $(y_1,\ldots,y_T)$. \textcolor{black}{The length of summary $T$ is controlled 
in accordance with the prototype length $K$.
} 
\end{task}

\begin{figure}[t]
\centering
 \includegraphics[width=8.3cm]{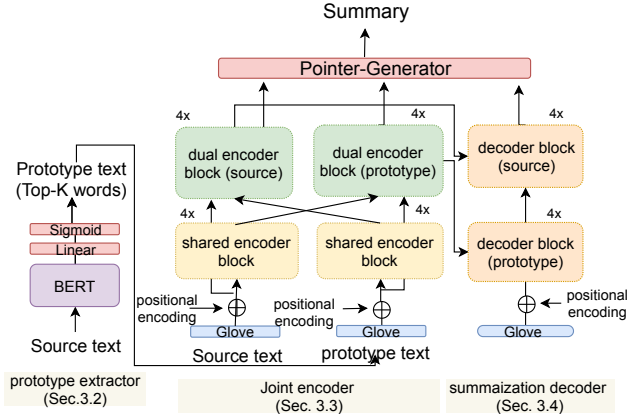}
  \caption{Architecture of proposed model\label{fig:proposed_model}}
\end{figure}

\section{Proposed Model}
\subsection{Overview}
Our model consists of three modules: the prototype extractor, joint encoder, and summary decoder (Figure~\ref{fig:proposed_model}). 
\textcolor{black}{The last two modules comprise Task 2, the prototype-guided}
\textcolor{black}{abstractive summarization}.
The prototype extractor uses BERT, while the joint encoder and summary decoder use the Transformer architecture~\cite{transformer}. 

\subsubsection{Prototype extractor (\S\ref{sec:prototype_extractor})} The prototype extractor extracts the top-$K$ important words from the source text. 
\subsubsection{Joint encoder (\S\ref{sec:joint_encoder})}
{The joint encoder} encodes both the source text and the prototype text. 
\subsubsection{Summary decoder (\S\ref{sec:summary_decoder})}
The summary decoder is based on the pointer-generator model and generates an abstractive summary by using the output of the joint encoder.

\subsection{Prototype Extractor\label{sec:prototype_extractor}}
Since our model extracts the prototype at the word level, the prototype extractor estimates an \textcolor{black}{importance} score $p^{\rm ext}_l$ of each word $x^C_l \in X^C$. 
BERT has achieved SOTA on many classification tasks, so it is a natural choice for the prototype extractor. Our model uses BERT and a task-specific feed-forward network on top of BERT. 
We tokenize the source text using the BERT tokenizer{\footnote{\url{https://github.com/google-research/bert/}}} and fine-tune the BERT model.
The \textcolor{black}{importance} score $p_{l}^{\rm ext}$ is defined as 
\begin{equation}
\label{eq:sigmoid}
   p_l^{\rm ext}= \sigma(W_1^\top {\rm BERT}(X^C)_l+b_1) 
\end{equation}
where ${\rm BERT ()}$ is the last hidden state of the pre-trained BERT. $W_1 \in \mathcal{R}^{d_{bert}}$ and $b_1$ are learnable parameters. $\sigma$ is a  sigmoid function. $d_{bert}$ is the dimension of the last hidden state of the pre-trained BERT. 



\textcolor{black}{To extract a more fluent prototype than when using only the word-level importance, 
we define a new weighted importance score $p^{{\rm ext}_w}_l$ that 
incorporates a sentence-level importance score as a weight for the 
word-level importance score:}
\begin{eqnarray}
\label{eq:weighted_saliency_probability}
   p_l^{{\rm ext}_w} = p_l^{\rm ext} \cdot p^{\rm ext}_{S_j} , \hspace{1em}
p^{\rm ext}_{S_j} = \frac{1}{N_{S_j}}\sum_{l:x_l \in S_j}{p_l^{\rm ext}}  
\end{eqnarray}
where $N_{S_j}$ is the number of words in the $j$-th sentence $S_j \in X^C$. 
\textcolor{black}{Our model extracts the top-$K$ important words as a prototype from the source text on the basis of $p^{{\rm ext}_w}_l$. It controls the length of the summary in accordance with the number of words in the prototype text, $K$.}

\subsection{Joint Encoder}\label{sec:joint_encoder}

\subsubsection{Embedding layer}
This layer projects each of the one-hot 
vectors of words
$x^C_l$ (of size $V$) into a $d_{word}$-dimensional vector space with a pre-trained weight matrix $W^e \in \mathcal{R}^{d_{word} \times V}$ such as GloVe~\cite{glove}. Then, the 
word
embeddings are mapped to $d_{model}$-dimensional vectors by using the fully connected layer, and the mapped embeddings are passed to a ReLU function.
This layer also adds positional encoding to the word embedding~\cite{transformer}. 

\subsubsection{Transformer encoder blocks}
The 
encoder encodes the embedded source and prototype texts with a stack of Transformer blocks~\cite{transformer}.
%
Our model encodes 
the two texts
with the encoder stack independently. We denote these outputs as $E_s^C \in \mathcal{R}^{d_{model} \times L}$ and $E_s^P \in \mathcal{R}^{d_{model} \times K}$, respectively.  

\subsubsection{Transformer dual encoder blocks}
\textcolor{black}{This block} calculates the interactive alignment between the encoded source and prototype texts. Specifically, it encodes the source and prototype texts and then performs multi-head attention on the other output of the encoder stack (i.e., $E_s^C$ and $E_s^P$).
We denote the outputs of \textcolor{black}{the dual encoder stack} of the source text and prototype text by  $M^C \in \mathcal{R}^{d_{model} \times L}$ and  $M^P \in \mathcal{R}^{d_{model} \times K}$, respectively.

\subsection{Summary Decoder}\label{sec:summary_decoder}
\subsubsection{Embedding layer}
The 
decoder receives a sequence of words in an abstractive summary $Y$, which is generated through an auto-regressive process. At each decoding step $t$, this layer projects each of the one-hot vectors of the words $y_t$ in the same way as the embedding layer in the joint encoder.

\subsubsection{Transformer decoder blocks}
The 
decoder uses a stack of decoder Transformer blocks~\cite{transformer} that perform multi-head attention on the encoded representations of the prototype, $M^P$. It uses another stack of decoder Transformer blocks that perform multi-head attention on those of the source text, $M^C$, on top of the first stack. The first stack rewrites the prototype text, and the second one complements the rewritten prototype with the original source information. 
\textcolor{black}{The subsequent mask is used in the stacks since 
this component is used in a step-by-step manner at test time.}
The output of the stacks is $M^S \in \mathcal{R}^{d_{model} \times T}$.

\subsubsection{\textcolor{black}{Copying mechanism}}
Our pointer-generator model copies the words from the source and prototype texts \textcolor{black}{on the basis of the copy distributions}, for efficient reuse. 
\subsubsection{Copy distributions}
The copy distributions of the source and prototype words are described as follows:
\begin{eqnarray}
   \nonumber
   p_p(y_t) = \sum_{k:x_k^P=y_t} \alpha_{tk}^P, \hspace{1em}  
   p_c(y_t) = \sum_{l:x_l^C=y_t} \alpha_{tl}^C
\end{eqnarray}
where $\alpha_{tk}^P$ and $\alpha_{tl}^C$ are respectively the first attention heads \textcolor{black}{of the last block} in the first and second stacks of the 
decoder.

\subsubsection{Final vocabulary distribution}
The final vocabulary distribution is described as follows:
\begin{eqnarray}
\nonumber
   && p(y_t) = \lambda_g p_g(y_t) + \lambda_c p_c(y_t) + \lambda_p p_p(y_t)  \\
   && \lambda_g, \lambda_c, \lambda_p \ = {\rm softmax}(W^v[M_t^S;c_t^C;c_t^P]+ b^v) \nonumber \\
   && c_t^C = \sum_l \alpha_{tl}^C M_l^C 
   , \hspace{1em}
   c_t^P =  \sum_k \alpha_{tk}^P M_k^P \nonumber \\
   \nonumber && p_g(y_t)={\rm softmax}(W^{g}(M_t^S) + b^g)
\end{eqnarray}
where $W^v \in \mathcal{R}^{3\times 3d_{model}}$, $b^v \in \mathcal{R}^3$, $W^g \in \mathcal{R}^{d_{model} \times V}$, and $b^g \in \mathcal{R}^V$ are learnable parameters.

\section{Training}
\textcolor{black}{Our model is not trained in an end-to-end manner: the prototype extractor is trained first and then the encoder and decoder are trained.}
\subsection{Generating Training Data}
\subsubsection{
Prototype extractor\label{subsubsec:make_training_data}}
Since there are no supervised data for the prototype extractor, we created pseudo training data like in \cite{bottomup}. The training data consists of word $x^C_l$ and label $r_l$ pairs,  ($x^C_l$, $r_l$) for all $x^C_l \in X^C$. $r_l$ is 1 if $x^C_l$ is included in the summary; otherwise it is 0.
To construct the paired data automatically, we first extract oracle source sentences $S^{oracle}$ that maximize the ROUGE-R score in the same way as in \cite{unified}. Then, we calculate the word-by-word alignment between the reference summary and $S^{oracle}$ using a dynamic programming algorithm  \textcolor{black}{to consider the word order}. Finally, we label all aligned words with 1 and other words, \textcolor{black}{including the words that are not in the oracle sentence}, with 0.

\subsubsection{Joint encoder and summary decoder}
We have to create triple data of ($X^C$, $X^{P}$, $Y$), consisting of the source text, the gold prototype text,
and the target text, for training \textcolor{black}{our encoder and decoder}.
\textcolor{black}{
We use the top-$K$ words (in terms of $p^{{\rm ext}_w}_l$; Eq.~\ref{eq:weighted_saliency_probability}) in the oracle sentences $S^{oracle}$ as the gold prototype text 
to extract a prototype closer to the reference summary 
and improve the quality of the encoder-decoder training.
} 
$K$ is decided using the reference summary length $T$. 
To obtain a natural summary close to the desired length, we quantize the 
length $T$ into discrete bins, where each bin represents a size range. 
\textcolor{black}{We set the size range to 5 in this study. 
That is, the value nearest to the summary length $T$ among multiples of 5 is selected for $K$.}

\subsection{Loss Function}\label{subsec:loss}
\subsubsection{Prototype extractor}
We use the binary cross-entropy loss, because the 
extractor estimates the importance score of each word (Eq.~\ref{eq:sigmoid}), which is a 
binary classification task.
\begin{gather}
\nonumber
L_\mathrm{ext} = -  \frac{1}{NL} \sum_{n=1}^N \sum_{l=1}^L \biggl(
\begin{split}
&r_l \log p^\mathrm{ext}_l +  \\
&(1-r_l) \log (1-p^{\mathrm{ext}}_l) 
\end{split}
\biggr)
\end{gather}
where $N$ is the number of training examples.
\subsubsection{Joint encoder and summary decoder}
The main loss for the encoder-decoder is the cross-entropy loss: 
\begin{align}
\nonumber
L_{\rm gen}^{\rm main} &= -\frac{1}{NT}\sum_{n=1}^{N}\sum_{t=1}^{T} \log p(y_t|y_{1:t-1}, \textcolor{black}{X^C, X^P}).
\end{align}
Moreover, we add the attention guide loss of the summary decoder. This loss is designed to guide the estimated attention distribution to the reference attention.
\begin{align}
\nonumber
L_{\rm attn}^{\rm sum} &= -\frac{1}{NT}\sum_{n=1}^{N}\sum_{t=1}^{T} \log \alpha_{t,l{(t)}}^{C}  \\ \nonumber
L_{\rm attn}^{\rm proto} &= -\frac{1}{NT}\sum_{n=1}^{N}\sum_{t=1}^{T} \log \alpha_{t,l{(t)}}^{\rm proto} 
\end{align}
\textcolor{black}{$\alpha_{t,l{(t)}}^{\rm proto}$ is the first attention head of the last block in the joint encoder stack for the prototype.}
$l(t)$ denotes the absolute position in the source text corresponding to the $t$-th word in the sequence of summary words.
The overall loss of the generation model is a linear combination of these three losses.
\begin{align}
\nonumber
L_{\rm gen} = L_{\rm gen}^{\rm main} + \lambda_{1} L_{\rm attn}^{\rm sum} + \lambda_{2} L_{\rm attn}^{\rm proto}
\end{align}
$\lambda_1$ and $\lambda_2$ were set to 0.5 in the experiments. 

\section{Inference}
During the inference period, 
we use a beam search and re-ranking~\cite{sentence_rewriting}. 
We keep all $N_{\rm beam}$ summary candidates provided by the beam search, where $N_{\rm beam}$ is the size of the beam, and generate the $N_{\rm beam}$-best summaries. The summaries are then re-ranked by the number of repeated N-grams, the smaller the better. The beam search and this re-ranking 
improve the ROUGE score of the output, as they eliminate candidates that contain repetitions. For the length-controlled setting, we set the value of $K$ to the desired length. For the standard \textcolor{black}{setting}, we set it to the average length of the reference summary in the validation data.

\section{Experiments}
\subsection{Datasets and settings}
\subsubsection{Dataset}
We used the CNN/DM dataset~\cite{CNNDM}, a standard corpus for news summarization.
The summaries are bullet points for the articles shown on their respective
websites. Following \citet{see17}, we used the non-anonymized version of the corpus and truncated the source documents to 400 tokens and the target summaries to 120 tokens. The dataset includes 286,817 training pairs, 13,368 validation pairs, and 11,487 test pairs.
We also used the NEWSROOM dataset~\cite{NEWSROOM}. NEWSROOM contains various news sources (38 different news sites).
We used 973,042 pairs of data for training. We sampled 30,000 pairs for validation data, and the number of the test pairs was 106,349. 
\textcolor{black}{To evaluate the length-controlled setting for NEWSROOM dataset, we randomly sampled 10,000 samples from the test set.}

\subsubsection{Model Configurations}
We used the same configurations for the two datasets.
The extractor used the pre-trained BERT$_\textrm{large}$ model~\cite{BERT}.
We fine-tuned BERT for two epochs with the default settings. 
Our encoder and decoder
used pre-trained 300-dimensional GloVe embeddings. 
The encoder and decoder Transformer have four blocks.
The number of heads was 8, and the number of dimensions of 
FFN
was 2048. 
$d_{model}$ was set to 512.
We used the Adam optimizer~\cite{KingmaB15} with a scheduled learning rate \cite{transformer}.
We set the size of the input vocabulary to 100,000 and the output vocabulary to 1,000. 



\subsection{Evaluation Metrics}
\textcolor{black}{We used the ROUGE scores (F1), including ROUGE-1 (R-1), ROUGE-2 (R-2), and ROUGE-L (R-L), as the evaluation metrics~\cite{Lin:2004}. 
We used the files2rouge toolkit for calculating the ROUGE scores}\footnote{\url{https://github.com/pltrdy/files2rouge}}.

\subsection{Results}

\begin{table}[t]
\begin{center}
\scalebox{0.85}[0.85]{
\begin{tabular}{c|l|c|c|c} \hline
Length & Model & R-1  & R-2 & R-L \\ \hline
 & LC$^1$ & \bf 19.03 & 8.45 & 16.47 \\
10 & LenEmb & 18.19 & \bf 8.96 & \bf 17.44 \\
& LPAS & 17.43 & 8.87 &  16.78 \\ \hline
 & LC & 32.26 & 13.60 & 24.64 \\
30 & LenEmb & 34.01 & 15.51 & 31.43 \\
& LPAS & \bf 35.11 & \bf 17.21  & \bf 32.83 \\ \hline
 & LC & 34.71 & 14.24 & 25.62 \\
50 & LenEmb & 38.66 & 17.17 & 35.49 \\
& LPAS & \bf 41.47 & \bf 19.70 & \bf 38.46 \\ \hline
 & LC & 33.83 & 13.67 & 24.67 \\
70 & LenEmb & 39.57 & 17.38 & 36.22 \\
& LPAS & \bf 42.48 & \bf 19.97 & \bf 39.25 \\ \hline
 & LC & 32.17 & 13.00 & 23.28 \\
90 & LenEmb & 38.51 & 16.79 & 35.24 \\
& LPAS & \bf 41.54 & \bf 19.43 & \bf 38.30 \\ \hline\hline
 & LC & 30.40 & 12.59 & 22.94 \\
AVG & LenEmb & 33.79 & 15.16 & 31.16 \\
& LPAS & \bf 35.60 & \bf 17.04 & \bf 33.12 \\ \hline
\end{tabular}
}
\end{center}
\caption{ROUGE scores (F1) of abstractive summarization models with different lengths on the CNN/DM dataset (10, 30, 50, 70, 90 words).
AVG indicates the average ROUGE score for the five different lengths.
$^1$\citep{CNNlengthcontrol}
\label{tab:result_generation_length_control}
}
\end{table}

\subsubsection{Does our model improve the ROUGE score in the length-controlled setting?}
\textcolor{black}{We used two types of length-controllable models as baselines. The first one is a CNN-based length-controllable model (LC) that uses the desired length as an input to the initial state of the CNN-based decoder. ~\cite{CNNlengthcontrol}. The second one (LenEmb) embeds the remaining length and adds them to each decoder step~\cite{kikuchi16}. Since there are no previous results on applying LenEmb to the CNN/DM dataset, we implemented it 
as a Transformer-based encoder-decoder model. Specifically, we simply added the embeddings of the remaining length to the word embeddings at each decoding step.}

Table~\ref{tab:result_generation_length_control} shows 
that our model achieved high ROUGE scores for different lengths and outperformed the previous length-controllable models in most cases. Our model was about 2 points more accurate on average than LenEmb.
Our model selected the most important words from the source text in accordance with the \textcolor{black}{desired}
length. 
It was thus effective at keeping the important information even in the length-controlled setting. 
Figure~\ref{fig:rouge_length}a shows the precision, recall, and F score of ROUGE for different lengths. 
Our model maintained a high \textcolor{black}{F-score} around the average length (around 60 words); this indicates that it can select important information and generate stable results with different lengths.

\subsubsection{Does our model generate a summary with the desired length?}
\begin{figure}[t]
\centering
 \includegraphics[height=2.7cm]{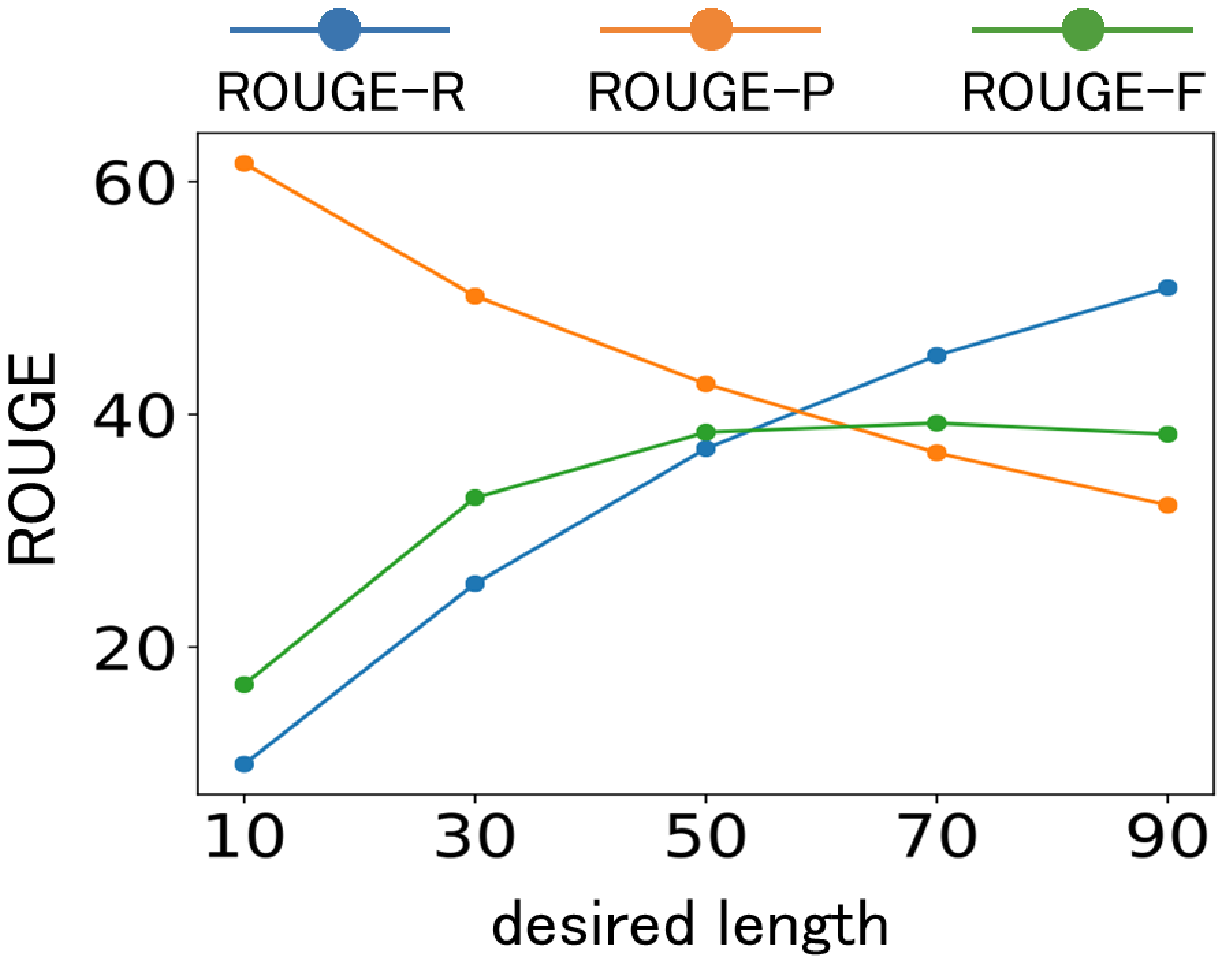}
 \includegraphics[height=2.8cm]{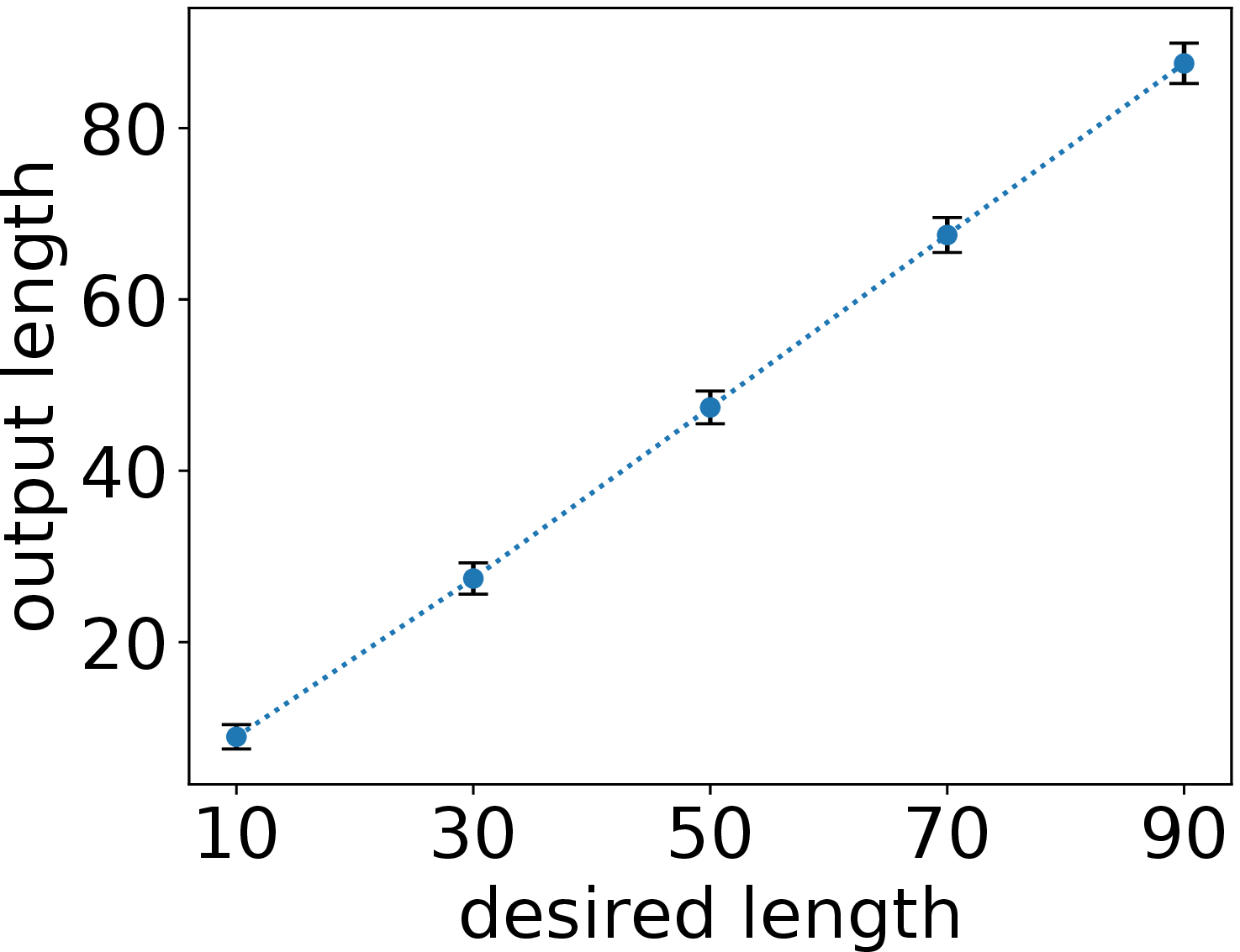}
    \caption{Results in the length-controlled setting on CNN/DM. a): ROUGE-L recall, precision and F scores for different lengths (left). 
  b): Output length distribution  (right).\label{fig:rouge_length}}
\end{figure}

Figure~\ref{fig:rouge_length}b shows the relationship between the desired length and the output length. The x-axis indicates the desired length, and the y-axis indicates the average length and standard deviation of the length-controlled output summary. The results show that our model properly controls the summary length. 
This controllable nature comes from the training procedure. When training \textcolor{black}{our encoder-decoder},  we set the number of words $K$ in the prototype text according to the length of the reference summary; therefore, the model learns to generate a summary that has a similar length to 
the prototype text.

\subsubsection{\textcolor{black}{How good is the quality of the prototype text?}}
To evaluate the quality of the prototype,
we evaluated the ROUGE scores of the extracted prototype text. 
Table~\ref{tab:result_content_selection} shows the results. 
In the table, LPAS-ext (top-3 sents) means the top-three sentences were extracted using $p^{\rm ext}_{S_j}$. 
\textcolor{black}{Interestingly, 
ROUGE-1 and ROUGE-2 scores of the LPAS-ext (Top-$K$ words) were higher than those of the sentence-level extractive models.} 
This indicates that word-level LPAS-ext is effective at finding not only important words (ROUGE-1), but also important phrases (ROUGE-2). 
Also, \textcolor{black}{we can see from Table~\ref{tab:result_generation_model} that whole LPAS improved} the ROUGE-L score of LPAS-ext. This indicates that our \textcolor{black}{joint encoder and summary decoder} generate more fluent summaries 
\textcolor{black}{with the help of}
the prototype text.
\begin{table}[t]
\begin{center}
\scalebox{0.90}[0.90]{
\begin{tabular}{l|c|c|c} \hline
 & R-1  & R-2  &  R-L \\ \hline
Lead3 & 40.3 & 17.7 & 36.6 \\
Bottom-Up (top-3 sents)$^1$ & 40.7 & 18.0 & 37.0 \\
Bottom-Up (word)$^1$ & 42.0 & 15.9 & 37.3 \\
NeuSum$^2$ & 41.6 & 19.0 & 38.0 \\ 
BertSum$^3$ & 43.25 & 20.24 & {\bf 39.63} \\ 
HIBERT$^4$ & 42.37 & 19.95 &  38.83 \\ \hline
LPAS-ext & & & \\ 
 - top-3 sents & 41.48 & 19.23 & 37.76 \\ 
 - Top-$K$ words & {\bf 44.79} & {\bf 20.59} & 38.12 \\ \hline
\end{tabular}
}
\end{center}
\caption{ROUGE scores (F1) of our prototype extractor (LPAS-ext) on CNN/DM\label{tab:result_content_selection}. $^1$\citep{bottomup}; $^2$\citep{score_and_select}; $^3$\citep{Bertsum}; $^4$\citep{HIBERT}}
\end{table}

\subsubsection{Does our abstractive model improve if the quality of the prototype is improved?}
\begin{table}[t]
\begin{center}
\scalebox{0.95}[0.95]{
\begin{tabular}{l|c|c|c} \hline
 & R-1  & R-2  &  R-L \\ \hline
Average length &  42.55 & 20.09 & 39.36 \\ \hline
Gold length & 43.23 & 20.46 &  40.00 \\ 
Gold sentences + Gold length & 46.68 & 23.52 & 43.41 \\ \hline
\end{tabular}
}
\end{center}
\caption{ROUGE scores (F1) of abstractive summarization models with gold settings on the CNN/DM dataset.\label{tab:result_generation_model_goldlength}}
\end{table}
We evaluated our model in 
\textcolor{black}{the following two settings}
in order to analyze the relationship between the quality of the abstractive \textcolor{black}{summary} and that of the prototype. In the gold-length setting, we only gave the gold length \textcolor{black}{$K$} to \textcolor{black}{the prototype extractor}. In the gold sentences + the gold-length setting, we gave the gold sentences $S^{oracle}$ and gold length~(see \ref{subsubsec:make_training_data}). 
Table~\ref{tab:result_generation_model_goldlength} shows the results.
These results indicate that selecting the correct number of words in the prototype improves the ROUGE scores. In this study, we simply selected the average length when extracting the prototype for all examples \textcolor{black}{in the standard setting}; however, there 
\textcolor{black}{will}
be an improvement if we adaptively select the number of words in the prototype for each source text. Moreover, the ROUGE score largely improved in the gold sentence and gold-length settings. 
\textcolor{black}{This indicates that the quality of the generated summary will significantly improve by increasing the accuracy of the extractive model.}

\begin{table}[t]
\begin{center}
\scalebox{0.90}[0.90]{
\begin{tabular}{c|l|c|c|c} \hline
Length & Model & R-1  & R-2 & R-L \\ \hline
10 & LenEmb & \bf 22.99 & 13.42 & 21.45 \\
& LPAS & 22.80  & \bf 13.91  & \bf 21.59 \\ \hline
30 & LenEmb & 37.49 & 25.67 & 34.26  \\
& LPAS & \bf 39.22 & \bf 27.33  & \bf  35.95 \\ \hline
50 & LenEmb & 36.91 & 25.50 & 33.86 \\
& LPAS & \bf 38.57 & \bf 27.07 & \bf 35.44 \\ \hline
70 & LenEmb & 33.52 & 23.02 & 30.90 \\
& LPAS & \bf 35.29 & \bf 24.72 & \bf 32.62 \\ \hline
90 & LenEmb & 30.04 & 20.49 & 27.80 \\
& LPAS & \bf 31.53 & \bf 22.03 & \bf 29.30 \\ \hline\hline
AVG & LenEmb & 32.19 & 21.62 & 29.66  \\
& LPAS & \bf 33.48 & \bf 23.01 & \bf 30.98 \\ \hline
\end{tabular}
}
\end{center}
\caption{ROUGE scores (F1) of abstractive summarization models with different lengths on the NEWSROOM dataset. 
\label{tab:result_generation_length_control_newsroom}
}
\end{table}

\subsubsection{Is our model effective on other \textcolor{black}{datasets?}}
\begin{figure}[t]
\centering
 \includegraphics[height=2.9cm]{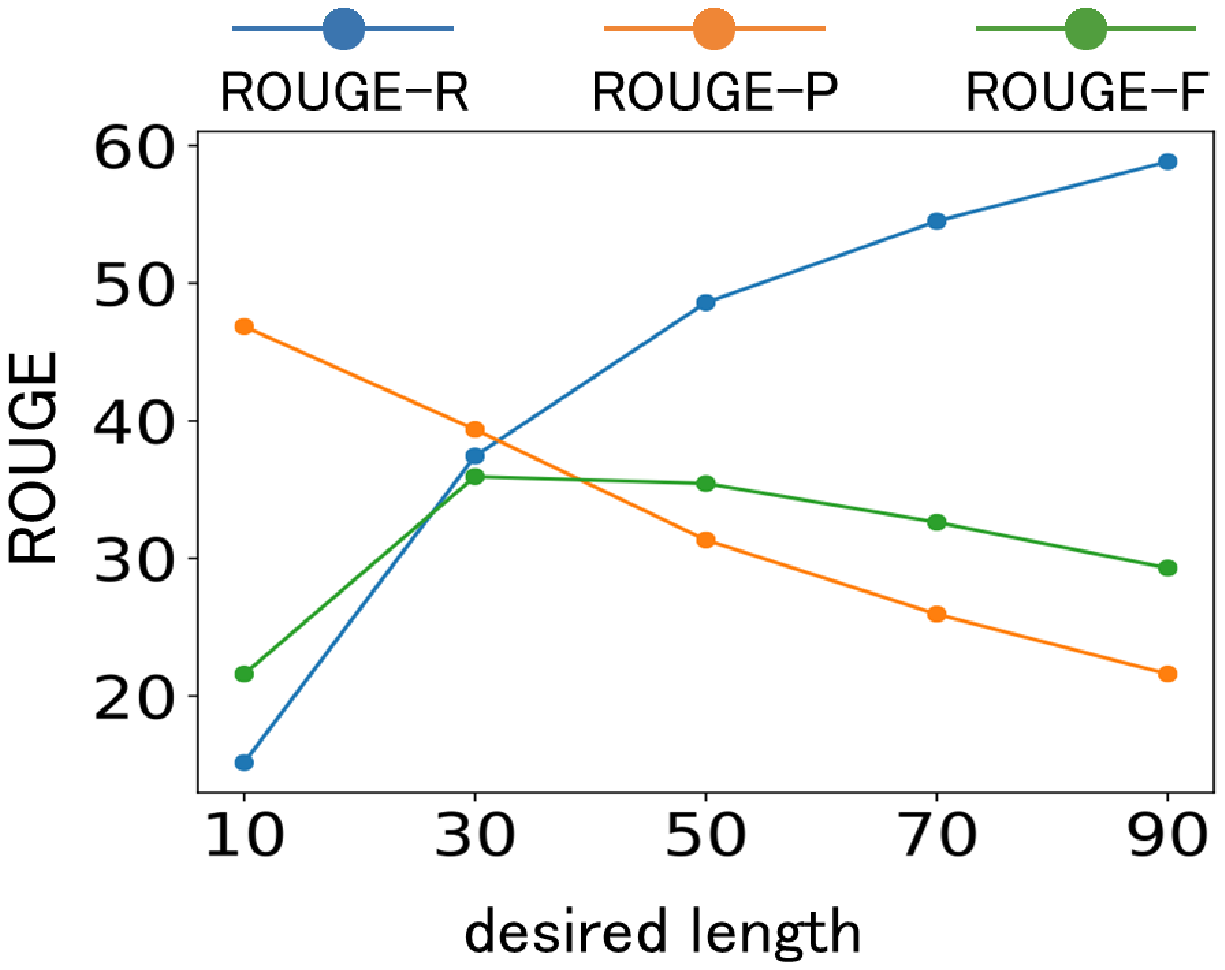}
 \includegraphics[height=2.9cm]{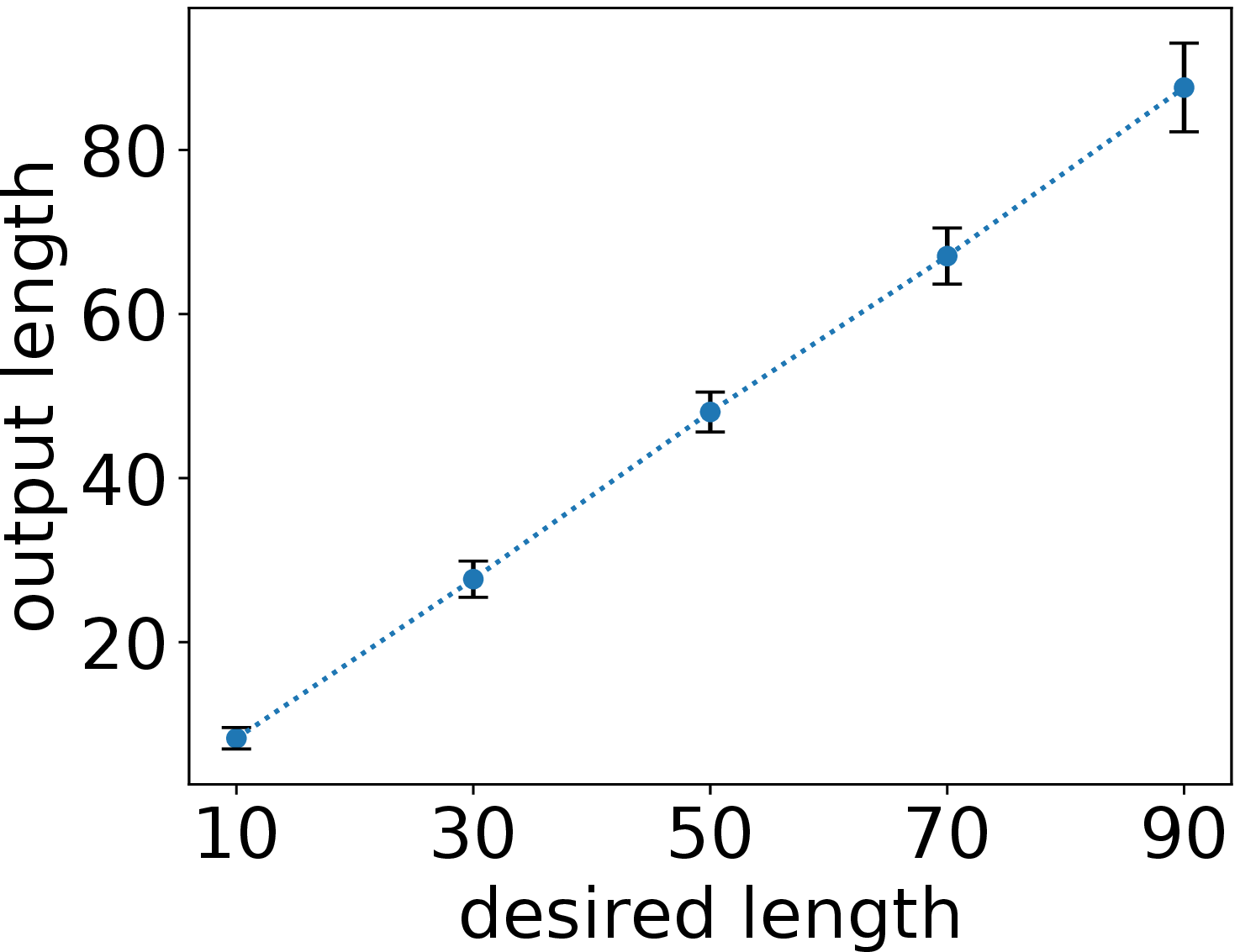}
   \caption{Results in the length-controlled setting on NEWSROOM. a): ROUGE-L recall, precision and F scores for different lengths (left). 
  b): Output length distribution (right).\label{fig:rouge_length_nr}}
\end{figure}

To verify the effectiveness of our model on various other summary styles, we evaluated it on a large and varied news summary dataset, NEWSROOM. 
 \textcolor{black}{
Table~\ref{tab:result_generation_length_control_newsroom} \textcolor{black}{and Figure~\ref{fig:rouge_length_nr} show} the results  
in the length-controlled setting for NEWSROOM. Our model achieved higher ROUGE scores than those of LenEmb.
 From Figure~\ref{fig:rouge_length_nr}a, we can see that the F-value of the ROUGE score was highest around 30 words. This is because the average word number is about 30 words. Moreover, Figure~\ref{fig:rouge_length_nr}b shows that our model also acquired a length control capability for a dataset with various styles.}

\begin{table}[t]
\begin{center}
\scalebox{0.85}[0.89]{
\begin{tabular}{l|c|c|c} \hline
 Model & R-1  & R-2 & R-L \\ \hline
\underline{w/o pre-trained encoder-decoder model} &  &  &  \\ 
Pointer-Generator$^1$ & 36.44 & 15.66 & 33.42 \\
Pointer-Generator + Coverage$^1$ & 39.53 & 17.28 & 36.38 \\
Key information guide network$^2$ & 38.95 & 17.12 & 35.68 \\
Unified summarization$^3$ & 40.68 & 17.97 & 37.13 \\
Sentence-rewriting$^4$ & 40.88 & 17.80 &  38.54 \\
Bottom-Up$^5$ & 41.22 & 18.68 &  38.34 \\
EXCONSUMM Compressive$^6$ & 40.9 & 18.0 &  37.4 \\
ETADS $^7$ & 41.75 & 19.01 &  38.89 \\
\hline
LPAS  & \bf 42.55 & \bf 20.09 & \bf 39.36 \\ 
\ \ w/o Prototype & 40.71 & 18.43 & 37.32 \\ 
\ \ w/o Source & 40.08 & 18.32 & 37.08 \\ \hline\hline
\underline{w/ pre-trained encoder-decoder model} &  &  &  \\ 
PoDA$^8$ & 41.87 & 19.27 & 38.54 \\ 
UniLM$^{9}$  &  43.47 &  20.30 &  40.63 \\ 
T5$^{10}$ & 43.52 & \bf 21.50 & 40.69 \\ 
BART$^{11}$ & \bf 44.16 & 21.28 & \bf 40.90 \\ \hline
\end{tabular}
}
\end{center}
\caption{ROUGE scores (F1) of abstractive summarization models on CNN/DM\label{tab:result_generation_model}. 
$^1$\citep{see17}; $^2$\citep{keyword_guided}; $^3$\citep{unified}; $^4$\cite{sentence_rewriting}; $^5$\citep{bottomup};
$^6$\citep{EXCONSUMM_Compressive};$^7$\citep{ETADS}; $^8$\citep{poda}; $^9$\citep{unilm}; $^{10}$\citep{2019t5}; $^{11}$\citep{BART}.
\textcolor{black}{LPAS} w/o Prototype denotes a simple Transformer-based pointer-generator, which is our 
model without the prototype extractor and the joint encoder. 
LPAS w/o Source denotes a model that generates a summary only from the prototype text.
}
\end{table}

\begin{table}[t]
\begin{center}
\scalebox{0.95}[0.95]{
\begin{tabular}{l|c|c|c} \hline
 & R-1  & R-2  &  R-L \\ \hline
Lead3 $^1$ & 32.02  &  21.08 & 29.59 \\ 
pointer-generator $^1$ & 27.54 & 13.32 & 23.50 \\ \hline
\textcolor{black}{LPAS}  &   &  &  \\
  $K$ = average length & 39.24  & 27.20 & 35.84 \\ 
  $K$ = domain length & {\bf 39.79} & {\bf 27.85} & {\bf 36.48} \\
  LPAS (w/o Prototype) & 38.48 & 26.99 & 35.30 \\
  \hline
\end{tabular}
}
\end{center}
\caption{ROUGE scores (F1) of proposed models on NEWSROOM dataset.\label{tab:result_generation_model_NEWSROOM} $^1$\citep{NEWSROOM}}
\end{table}

\subsubsection{\textcolor{black}{How \textcolor{black}{well does our model perform} in the standard setting?}}
Table~\ref{tab:result_generation_model} shows 
\textcolor{black}{that our model achieved the ROUGE scores comparable to previous models that do not consider the length constraint on the CNN/DM dataset.
We note that the current state-of-the-art models use pre-trained encoder-decoder models (8-11), while the encoder and decoder of our model (except for prototype extractor) were not pre-trained.
}


We also examined the results of generating a summary from only the prototype (LPAS w/o Source) \textcolor{black}{or the source (LPAS w/o Prototype)}. Here, using only the prototype, 
turned out to have the same accuracy as using only the source, but the model 
using the source and the prototype simultaneously had higher accuracy.
These results indicate that our prototype extraction and joint encoder effectively incorporated the source text and prototype information and contributed to improving the accuracy.

The results for the NEWSROOM dataset \textcolor{black}{under standard settings} are shown in Table~\ref{tab:result_generation_model_NEWSROOM}. To consider differences in summary length between news domains, we evaluated our model in the average length and domain-level average length (denoted as domain length) settings. 
\textcolor{black}{The results indicate that our model had significantly higher ROUGE scores compared with the official baselines and outperformed our baseline (LPAS w/o Prototype). 
}
They also indicate that our model is effective on datasets containing text in various styles. 
Moreover, we found that considering the domain length has positive effects on the ROUGE scores. This indicates that our model can easily reflect differences in summary length among various styles. 

\section{Related Work and Discussion}

\subsubsection{Length control for summarization}
\citet{kikuchi16} were the first to propose using length embedding for length-controlled abstractive summarization. \citet{controllable_summarization} also used length embeddings at the beginning of the decoder module for length control. \citet{CNNlengthcontrol} proposed a CNN-based length-controllable summarization model 
\textcolor{black}{that uses the desired length as an input to the initial state of the decoder.}
\textcolor{black}{\citet{positional_lc} introduced positional encoding that represents the remaining length at each decoder step of the Transformer-based encoder-decoder model. It is almost equivalent to the model LenEmb we implemented.}
These previous models use length embeddings for controlling the length in the decoding module, whereas we use the prototype extractor for controlling the summary length and to include important information in the summary.  

\subsubsection{Neural extractive-and-abstractive summarization}
\citet{unified}, \citet{bottomup} and \citet{ETADS} incorporated a sentence- and word-level extractive model in the pointer-generator model. Their models weight the copy probability for the source text by using an extractive model and guide the pointer-generator model to copy important words. \citet{keyword_guided} proposed a keyword guided abstractive summarization model. 
\citet{sentence_rewriting} proposed a sentence extraction and re-writing model that trains in an end-to-end manner by using reinforcement learning. \citet{retrieve_rewrite} proposed a search and rewrite model. \citet{EXCONSUMM_Compressive} proposed a combination of sentence-level extraction and compression.
The idea behind these models is word-level weighting for the entire source text or sentence-level re-writing. On the other hand, our model guides the summarization with a 
\textcolor{black}{length-controllable} prototype text by using the prototype extractor and joint encoder. 
Utilizing extractive results to control the length of the summary is a new idea.

\subsubsection{Large-scale pre-trained language model}
BERT~\cite{BERT} is a new pre-trained language model that uses bidirectional encoder representations from Transformer. BERT has performed well in many natural language understanding tasks such as the GLUE benchmarks~\cite{WangSMHLB18} 
and natural language inference~\cite{WilliamsNB18}. 
\citet{Bertsum} used BERT for their sentence-level extractive summarization model. \citet{HIBERT} trained a new pre-trained model that considers document-level information for sentence-level extractive summarization. 
We used BERT for the word-level prototype extractor and verified the effectiveness of using it in the word-level extractive module. 
\textcolor{black}{Several researchers have published pre-trained encoder-decoder models very recently~\cite{poda,BART,2019t5}. \citet{poda} pre-trained a transformer-based pointer-generator model. \citet{BART} pre-trained a normal transformer-based encoder-decoder model using large unlabeled data and achieved state-of-the-art results. \citet{unilm} extended the BERT structure to handle sequence-to-sequence tasks. }

\subsubsection{Reinforcement learning for summarization}
Reinforcement learning (RL) is a key summarization technique. RL can be used to optimize non-differential metrics or multiple non-differential networks. \textcolor{black}{\citet{narayan-etal-2018-ranking} and \citet{dong-etal-2018-banditsum} used RL for extractive summarization.} \textcolor{black}{For abstractive summarization,} \citet{abstractive_RL} used RL to mitigate the exposure bias of abstractive summarization. \citet{sentence_rewriting} used RL to combine sentence-extraction and pointer-generator models.  
Our model achieved high ROUGE scores without RL. In future, we may incorporate RL in our models to get a further improvement. 

\section{Conclusion}
We proposed a new length-controllable abstractive summarization model. Our model consists of a word-level prototype extractor and a prototype-guided abstractive summarization model.  
The prototype extractor identifies the important part of the source text within the length constraint, and the abstractive model 
\textcolor{black}{is guided with the prototype text}.
This characteristic enabled it to achieve a high ROUGE score in standard summarization tasks. Moreover, our prototype extractor ensures the summary will have the desired length. \textcolor{black}{Experiments with the 
CNN/DM dataset and the NEWSROOM dataset show that our model outperformed previous models in standard and length-controlled settings}. In future, we would like to incorporate a pre-trained language model in the abstractive model to build a higher quality summarization model.

\bibliography{emnlp-ijcnlp-2019}
\bibliographystyle{aaai}

\end{document}